\documentclass[conference]{IEEEtran}
\usepackage{multirow}
\usepackage{amsmath}
\usepackage{graphicx}
\usepackage{subfig}
\usepackage{hyperref}
\usepackage{cleveref}
\graphicspath{ {pics/} }
\begin{document}

\title{Camera Model Identification Using Convolutional Neural Networks}

\author{\IEEEauthorblockN{Artur Kuzin}
\IEEEauthorblockA{Lead Data Scientist, Dbrain\\
ODS.ai\\
Moscow Institute of Physics and Technology\\
kuzin.artur@gmail.com}
\and
\IEEEauthorblockN{Artur Fattakhov}
\IEEEauthorblockA{Department of Innovations and High Technology\\
Moscow Institute of Physics and Technology\\
fattahov.ao@phystech.edu}
\and
\IEEEauthorblockN{Ilya Kibardin}
\IEEEauthorblockA{
Department of Innovations and High Technology\\
Moscow Institute of Physics and Technology\\
kibardin.ia@phystech.edu}
\and
\IEEEauthorblockN{Vladimir I.~Iglovikov}
\IEEEauthorblockA{Level 5, Autonomous Vehicle Division,\\ Lyft Inc\\
iglovikov@gmail.com}
\and
\IEEEauthorblockN{Ruslan Dautov}
\IEEEauthorblockA{
Big Data Institution\\
Shenzhen University\\
dautovri@szu.edu.cn}}


%
\maketitle

\begin{abstract}
Source camera identification is the process of determining which camera or model has been used to capture an image. In recent years, there has been a rapid growth of research interest in the domain of forensics. In the current work, we describe our Deep Learning approach to the camera detection task of 10 cameras as a part of the Camera Model Identification Challenge hosted by Kaggle.com where our team finished 2nd out of 582 teams with the accuracy on the unseen data of 98\%. Augmentations that allowed  a stay robust against  transformations. A number of experiments are carried out on datasets collected by organizers and scraped from the web.

\end{abstract}

\IEEEpeerreviewmaketitle
\section{Introduction}
In this work, we describe our solution for the IEEE's Signal Processing Society - Camera Model Identification Challenge hosted by Kaggle.com. Our team ranked the second place out of 582 teams in the task to create an algorithm that identifies the type of the camera that was used to capture an image. In general, the camera detection algorithm enables to solve many problems in forensics, such as identifies the owner of illegal or controversial materials (pedo-pornographic shots, terrorist act scenes, images that do not respect privacy laws, etc.), as well as helping to claim the intellectual property. 

There are two main ways to identify the type of camera. The first one uses metadata (e.g., EXIF tag) that keeps information about the type of the camera, and parameters used when the image was captured. The problem with this approach is that it is very unreliable since metadata can be easily manipulated or is unavailable if an image was re-saved or re-compressed.

The second approach based on the low-level features that are camera model specific and originate from processing steps carried within a camera. Every camera maker develops a set of sophisticated, non-linear algorithms that are applied to the raw image before saving it to the memory card. Examples include demosaicing, noise filtering, fixing lens distortion, etc.

Several camera identification algorithms were proposed in the literature, each trying to extract features related to different post-processing techniques. Some of them aim to extract features that are trying to exploit some apriori knowledge about imaging model, noise characteristics, demosaicing strategies, lens distortion, histogram strategies, etc. The others capture the statistical image properties and feed them to the machine learning classifiers. All of these approaches are not scalable in a sense that adding a new model of the camera will take a much manual effort to define the ways of feature extraction designated to track the traces.

In recent years, solutions based on Deep Learning techniques became mainstream in computer vision. Problems in satellite \cite{iglovikov2017satellite, Iglovikov_2018_CVPR_Workshops,
Buslaev_2018_CVPR_Workshops,
Seferbekov_2018_CVPR_Workshops}, medical \cite{shvets2018automatic,
rakhlin2018deep, shvets2018angiodysplasia, iglovikov2017pediatric} or any other imagery are successfully tackled with these techniques, routinely beating human performance in many tasks, including classification, segmentation, and detection. There are a few reasons why these techniques are popular. First, they are scalable in a way, that extending the model to work with new models is a straightforward task that does not require any special forensic domain knowledge or training, making this problem engineering rather than a scientific problem. Second, empirical evidence shows that the accuracy of the models is growing when more train data provided, which allows taking advantage of the fact that tremendous amounts of videos and images published on the internet.

There are many works where authors applied Deep Learning techniques to forensics. For example, in the paper \cite{barni2017aligned}, authors used CNNs to detect double JPEG Compression, while in \cite{bondi2017first}, authors used CNN, to extract features and SVM classifiers on top of extracted features.

The approach that we  propose follow a similar path, but we had a few essential modifications:
\begin{enumerate}
    \item We trained a network that performs predictions in an end2end manner.
    \item We used deep 161 Layer DenseNet architecture \cite{huang2017densely} that allows constructing very abstract representations of the low-level features created by the processing algorithms.
    \item We use our weight initialization model that was pre-trained on the ImageNet.
    \item We used aggressive data augmentations that allowed a model to stay robust against Gamma, Resize, Contrast and Resize transformations.
\end{enumerate}

This paper organized as follows. Section II reviews camera model identification papers, then, Section III describes the algorithm used in this work.  Section IV summarizes the lesson learned and experimental results.

 \section{Related work}
In the past decades, many methods have been proposed to determine which camera was used to take the image.

Most of the existing approaches that do not take metadata into account can be divided into two main categories: hardware and software source camera identification. Hardware category considers features of camera hardware, such as lens \cite{Wong2006, Dirik2008} and Charge Coupled Device (CCD) sensors \cite{Geradts2000}. However, software approach works with color filter array (CFA) interpolation artifacts \cite{Bayram2005, Yangjing2007, Celiktutan2006} and  Sensor Pattern Noise(SPN) \cite{lukas2006digital}.  

 Current promising approaches according to Van Lanh T et al \cite{VanLanh2007} are:

 \begin{itemize}
 \item Lens characteristics  \cite{Wong2006}
 \item Noise pattern in digital cameras \cite{lukas2006digital}
 \end{itemize}

 The first approach was proposed in the paper published by Choi et al. \cite{Wong2006} in 2006.  
 The method is focused on lens radial distortion where parameters of this distortion are used as features for the classification algorithm. This optical deviation occurs because of the use of the low-quality wide angle lenses which have a low cost.
  Manufacturers are implementing different lens systems to compensate for the radial distortion, where they are affecting the pattern of radial distortion.
 The critical limitation of the method is manual zooming or changing to custom lenses which decrease the accuracy of classification. In 2007, Van Lanh T et al. \cite{Van2007}
extended this approach and applied it to mobile phone cameras. This is one of the few methods that obtain results on early detection stage such as lenses.
 
The second approach which was initially proposed by Luka et al. \cite{lukas2006digital} in 2006.
Silicon wafers that are used during the production of the sensors have defects and different homogeneity. As a result, pixels at different positions have a different sensitivity to light which leads to a unique to each camera pattern of noise which is considered as the main component of Pixel Non-Uniformity(PRNU).

In \cite{Kang2012, Li2012} authors enhanced the prior algorithm by subtracting the average whitened sensor pattern noise. The limitation of this approach is the recommendation to use the smooth content images to extract relevantly reliable noise-based fingerprint \cite{Fridrich2009}.

In \cite{Kulkarni2015}, authors use a feature extraction pipeline, consisting of edge extraction using canny and Laplace operators, and combining them with the original images to extract Homogeneity, Contrast, Entropy, and Correlation. They used SVM and other classifiers on top of these features to obtain high accuracy results on the Dresden Dataset.

In the last few years, deep learning techniques were also applied to the camera detection task \cite{Bondi2017, Tuama2016, B2017}. Deep Learning approach has the advantage of working with extremely high capacity models, having tens of millions of free parameters. The power of the method is that Neural Networks do not require manual feature extraction as the model is learning the appropriate features directly from the data. It makes this method scalable in two ways. First of all, you can easily extend your detection algorithm to a big set of cameras, adding new models if needed. Second, the quality of the extracted features grows with the amount of the data that used for training. 

\subsection{Dataset}
Typically, camera identification algorithms evaluated on a Dresden Image Dataset \cite{Gloe:2010aa}. This dataset contains images from 74 cameras of 27 models with different scenes for each device (e.g., office, nature, etc.). However, it lacks augmentation and mobile phone cameras images.

In the current work, we used two datasets to evaluate our model performance. The first one was a dataset that was provided by the Organizers of the IEEE's Signal Processing Camera identification Challenge, and had of 2500 images, corresponding to ten camera models with 250 pictures each. Lens aberration proved to be a powerful feature in the previous work \cite{Wong2006}. To prevent the participants from using it, the organizers of the competition cropped central 500x500 parts of the images in the test set. Furthermore, half photos were augmented by the transformations Resize, Gamma, Contrast, or Jpeg Compression and the other half was in their raw form. For this dataset, the ground truth labels were unknown to the challenge participants, and evaluation was performed via LeaderBoard on the Kaggle.com website.

\begin{figure*}[t]
\centering
\begin{tabular}{cccccc}
\subfloat{\includegraphics[width=0.15\linewidth]{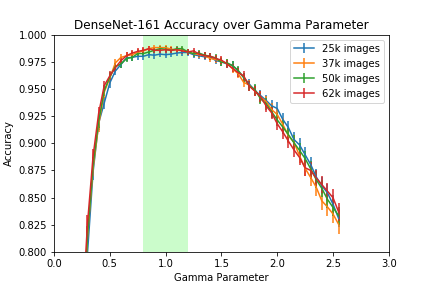}}&
\subfloat{\includegraphics[width=0.15\linewidth]{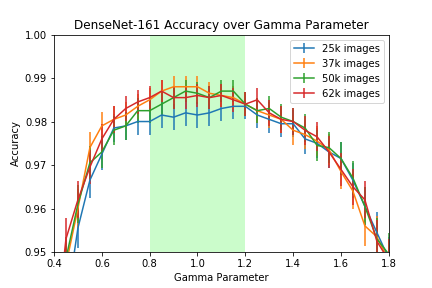}}&
\subfloat{\includegraphics[width=0.15\linewidth]{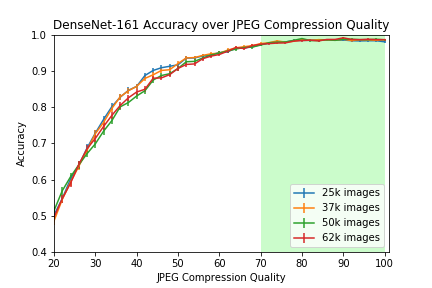}}&
\subfloat{\includegraphics[width=0.15\linewidth]{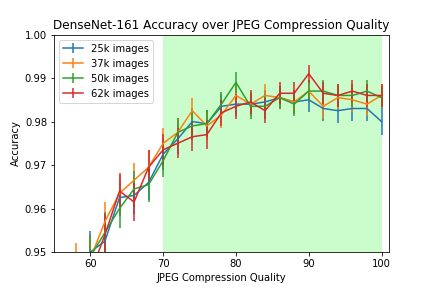}}&
\subfloat{\includegraphics[width=0.15\linewidth]{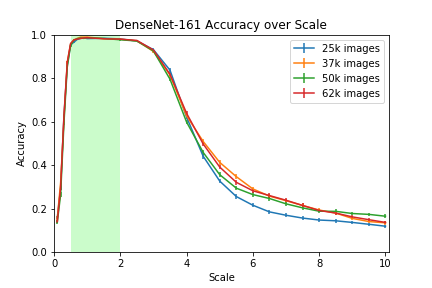}} &
\subfloat{\includegraphics[width=0.15\linewidth]{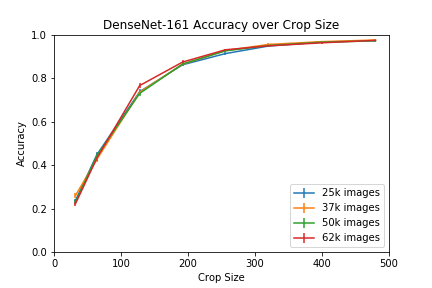}}\\
\subfloat{\includegraphics[width=0.15\linewidth]{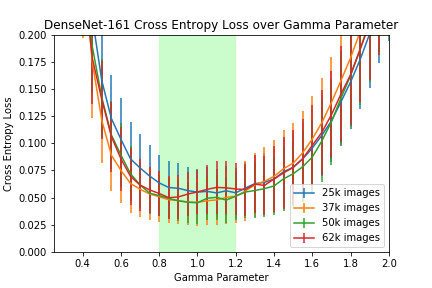}}&
\subfloat{\includegraphics[width=0.15\linewidth]{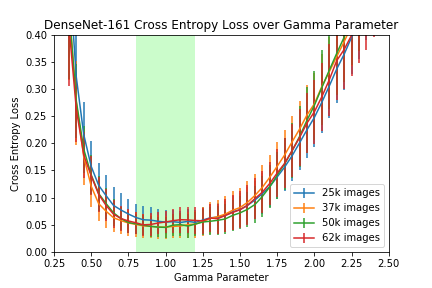}}&
\subfloat{\includegraphics[width=0.15\linewidth]{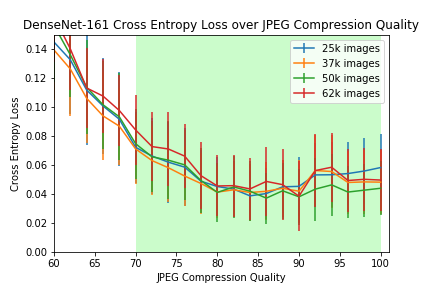}}&
\subfloat{\includegraphics[width=0.15\linewidth]{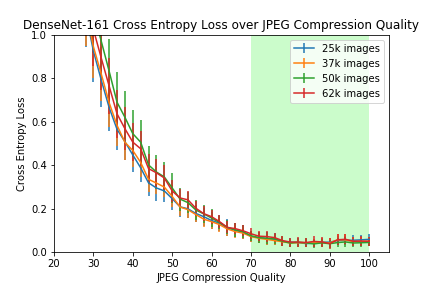}}&
\subfloat{\includegraphics[width=0.15\linewidth]{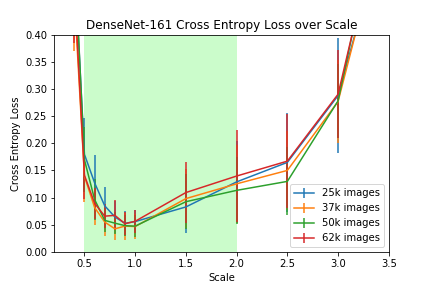}}&
\subfloat{\includegraphics[width=0.15\linewidth]{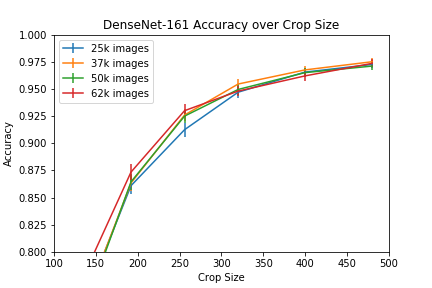}}\\
\end{tabular}
\caption{Accuracy and cross entropy loss over different transformations on DenseNet161. For each augmentation used 25k,50 and 62.3K training sizes. Ranges of parameters highlighted on graphs.}
\label{fig:results}
\end{figure*}

\begin{figure}[!hb]
\includegraphics[width=0.49\textwidth]{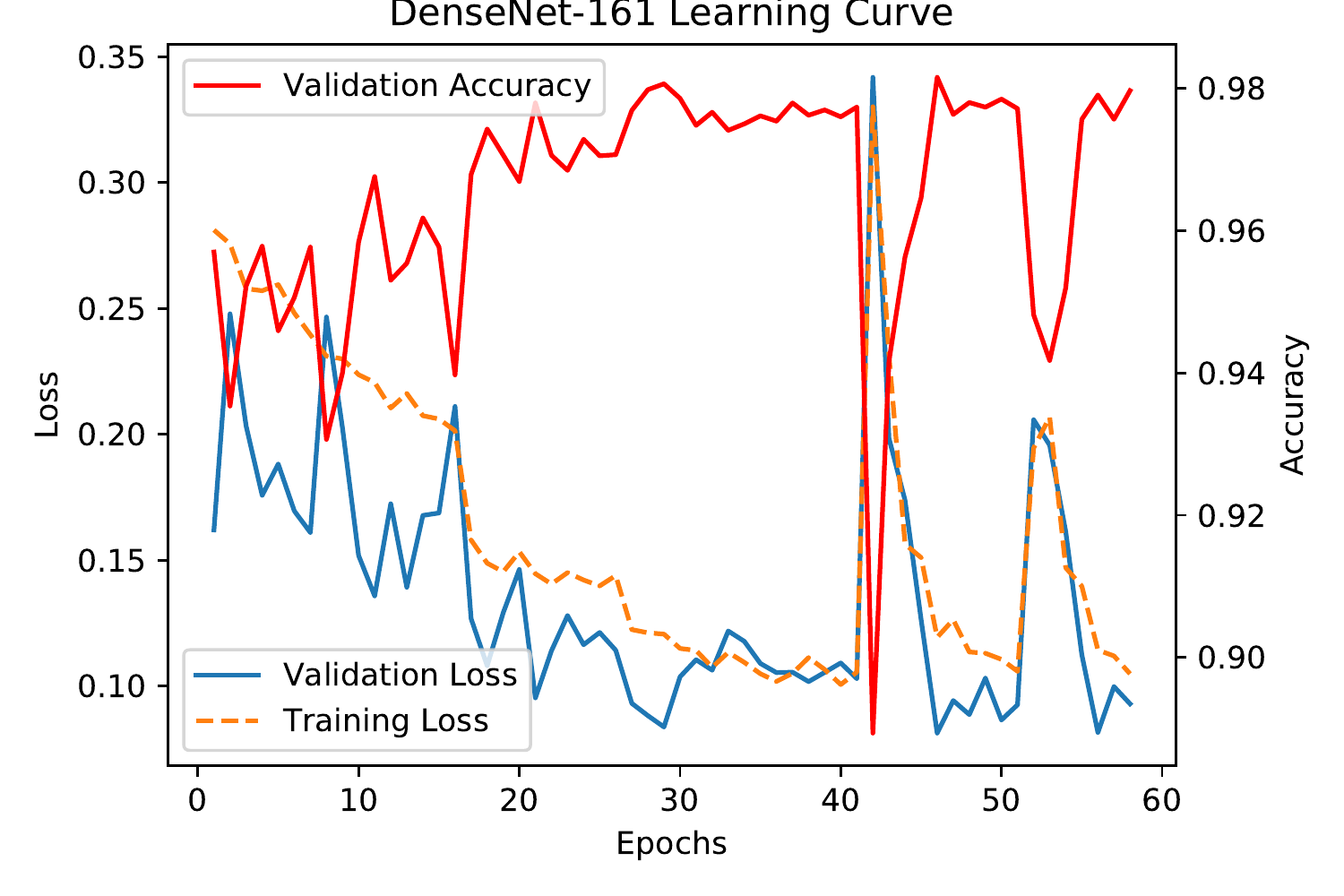}
\caption{Loss curve of network traning. Adam optimizer and 1e-3 as initial learning rate is used. With 2 GPUs (Nvidia 1080Ti) our implementation completes in about 2 days.}
\label{fig:loss}
\end{figure}

In the competition, external data was allowed to use. Hence we scrapped more than 500 Gb from Flickr, Yandex.Fotki, Wikipedia Commons, and mobile reviews websites to obtain images for the required ten classes. After this, we performed filtering based on the EXIF metadata, removing those that were manipulated by a Photoshop or LightRoom software. After this, images with Jpeg compression quality less than 95, were excluded. Finally, we filtered out images that had sizes that did not belong to the default list of possible image sizes that corresponding cameras generate. After this filtering, we got 78807 not-manipulated images. We split them into two parts: the train set(Table \ref{tab:dataset}) and 100 for the validation set that was used to evaluate our model performance locally and to perform an ablation study.

 \begin{table}[ht]
 \centering
  \begin{tabular}{ | l | l | l | }
        \hline
      \multirow{2}{*}{Camera model} &
      \multicolumn{2}{ |c| }{Subset} \\
      \cline{2-3}   
      & Training & Validation \\ \hline
      HTC-1-M7 & 10156 & 100 \\ \hline
      iPhone-6 & 10053 & 100 \\ \hline
      Motorola Droid Maxx & 11608 & 100 \\ \hline
      Motorola X & 1769 & 100 \\ \hline
      Samsung Galaxy S4 & 9351 & 100 \\ \hline
      iPhone 4S & 9383 & 100 \\ \hline
      LG Nexus 5X & 5437 & 100 \\ \hline
      Motorola Nexus 6 & 10950 & 100 \\ \hline
      Samsung Galaxy Note 3 & 6025 & 100 \\ \hline
      Sony NEX 7 & 3075 & 100 \\ \hline
  \end{tabular}
  \caption{Camera model classes with number of samples each part of the dataset. Table presents the final dataset which contains external and organizers datasets.}
  \label{tab:dataset}
\end{table}

\section{Algorithm}
Neural Networks are a universal approximator, that can learn any function from data, assuming that we have appropriate network architecture, enough training data, and proper training procedure. In the Camera Identification challenge at Kaggle, the organizers did not limit the use of the external data as it is typically happening in computer vision challenges. Because of this freedom we choose not to focus on the pre-processing steps but invest time into selecting the proper network architecture and training procedure.

For the network, we choose DenseNet 161 \cite{huang2017densely} which consists of the repeating convolutional blocks with an average pooling at the end before the last Dense layer that used for the final classification. The name DenseNet comes from that fact that skip connections are added between each pair of layers, which is believed to make the loss surface more smoothed, the optimization procedure not to get stuck in the local minimums. The GlobalAverage Pooling procedure before the final Dense layer is agnostic to the size in the XY dimension which allows using images of different sizes as an input.

During the competition, participants used different types of networks like ResNet\cite{2015arXiv151203385H}, ResNext\cite{2016arXiv161105431X}, DPN\cite{2017arXiv170701629C}, VGG\cite{2014arXiv1409.1556S}. All top teams used networks that have a large capacity with millions of the free parameters and were pre-trained on the ImageNet. In our experiments, DenseNet showed the best result, but we believe that as long as the architecture shows the good result on the ImageNet, transfer learning from it will show good result in the camera detection task.

During training, we randomly cropped patches of the size 960x960 and augmented them with the following transformations and applied 480x480 crops after this.

\begin{enumerate}
    \item Dihedral Group D4 transformations: Rotations by 90, 180, 270 degrees and flips.
    \item Gamma transformation. We choose the gamma parameter uniformly from the [0.8, 1.2] range.
    \item JPEG Compression with the parameters from 70 to 90.
    \item Scale transformations with the parameters sampled from the [0.5, 2] range.
\end{enumerate}

For all the above transformations we used an implementation from the albumentations \cite{buslaev2018albumentations} library.

After these transformations, images were collected into batches of the size 480x480 and used to train the network. As an optimizer we standard for classification problems cross entropy Loss:
\begin{equation}
    H_{y'}(y) := - \sum_{i} ({y_i' \log(y_i) + (1-y_i') \log (1-y_i)})
\end{equation}
We trained the network for 100k epochs using an Adam optimizer, with the initial learning rate as 1e-3. Loss curve is shown in Fig \ref{fig:loss}.

The test during the competition, we performed an inference on the 480x480 corner and center crops from the image, applying D4 transformation to each crop. All the above predictions were averaged. Our result with the score 0.987976 was the second out of 582 teams. This competition is evaluated on the weighted categorization accuracy: 
\begin{equation}
    \text{weighted accuracy}(y, \hat{y}) = \frac{1}{n} \sum_{i=1}^{n} \frac{w_i  (y_i = \hat{y}_i)}{\sum{w_i}}
\end{equation}
where $w_i$ is equal to 0.7 for unaltered images and 0.3 for altered images.

\section{Ablation study}

In addition to the participation in the challenge, we evaluated how the accuracy of our model is affected by the JPEG, gamma, and transformations.

First, we evaluated the dependence of the model quality from the training set size. We used 25k, 50k, and 62.5k. We did not find the statistical difference for these sizes. We believe that this counter intuitive result is related to the fact that for the chosen powerful architecture with the corresponding training schedule, this task was not challenging enough, which lead to a robust model with the validation accuracy of 0.98 on the smallest data point of  25k images. We believe that we needed to perform classification, not on ten but a much larger number of classes, say 100 or larger classes. The positive correlation between the size of the train data and model accuracy was more evident. 

Secondly, we evaluated the effect of the JPEG Compression, Gamma, and Resize augmentations on the validation accuracy. As shown Fig \ref{fig:results}. As expected our model consistently shows excellent performance in the ranges of the augmentations that were used during training. We believe that this result gives additional evidence that Deep Learning models can be made robust to a broad range of different transformations if desired transformations were used as a training time augmentations.

Finally, we estimated the effect of the crop size on the model performance. It is believed in the literature that algorithms that were used to process the raw images, leave low-level local features that can be used by the camera detection algorithms. We would assume that for the data that follow this assumption crop size, would not affect the model performance for a wide range of crop sizes. However, the curve Fig \ref{fig:results} may be interpreted as the fact that not just local, but long-range correlations between pixel values may serve as a powerful feature.

\section{Conclusion}

In the current work, we showed how the application of the deep learning techniques trained on the large amounts of the data. Data which scraped from the internet. In condition, good training schedule, network architecture and image augmentations could lead to a model that shows excellent performance in the camera detection task. Based on the proposed model, this paper studies can be straightforwardly applied in practice. In our work, we performed data filtering to avoid using manipulated images during training that significantly decreased the available imagery data. We believe that larger amounts of the `dirty` data used for training may result in a better quality model, but we did not perform a direct comparison of these two approaches in a current work leaving it for future research.

\section*{Acknowledgement}

The authors would like to thank Open Data Science community \cite{ODS} for many valuable discussions and educational
help in the growing field of machine/deep learning.

\bibliographystyle{IEEEtran}

\bibliography{lib}

\end{document}